\documentclass{article}
\pdfoutput=1 

\usepackage{PRIMEarxiv}

\usepackage[utf8]{inputenc} 
\usepackage[T1]{fontenc}    
\usepackage{hyperref}       
\usepackage{url}            
\usepackage{booktabs}       
\usepackage{amsfonts}       
\usepackage{nicefrac}       
\usepackage{microtype}      
\usepackage{fancyhdr}       
\usepackage{graphicx}       
\graphicspath{{media/}}     

\usepackage{amsmath}
\usepackage{amssymb}
\usepackage{bbm}
\usepackage{algorithm}
\usepackage{algorithmic}

\usepackage[backend=bibtex]{biblatex}
\title{Adaptive Resampling with Bootstrap for Noisy Multi-Objective Optimization Problems
}

\author{
  Timo Budszuhn \\
  Technische Universität Dortmund \\
  \texttt{timo.budszuhn@tu-dortmund.de} \\
   \And
 Mark Joachim Krallmann \\
  Technische Universität Dortmund \\
  \texttt{joachim.krallmann@tu-dortmund.de} \\
   \And
  Daniel Horn \\
  Technische Universität Dortmund \\
  \texttt{dhorn@statistik.tu-dortmund.de} 
}

\begin{document}
\maketitle

\begin{abstract}
The challenge of noisy multi-objective optimization lies in the constant trade-off between exploring new decision points and improving the precision of known points through resampling.
This decision should take into account both the variability of the objective functions and the current estimate of a point in relation to the Pareto front. Since the amount and distribution of noise are generally unknown, it is desirable for a decision function to be highly adaptive to the properties of the optimization problem. This paper presents a resampling decision function that incorporates the stochastic nature of the optimization problem by using bootstrapping and the probability of dominance. 
The distribution-free estimation of the probability of dominance is achieved using bootstrap estimates of the means. To make the procedure applicable even with very few observations, we transfer the distribution observed at other decision points. The efficiency of this resampling approach is demonstrated by applying it in the NSGA-II algorithm with a sequential resampling procedure under multiple noise variations.
\end{abstract}

\section{Introduction}
In the field of machine learning, models are becoming better but also more complex. Tuning hyperparameters while considering resources, costs, and performance forms a multi-objective optimization problem. Since performance evaluations are based on random processes, they are contaminated with noise. One of the open challenges in hyperparameter optimization is the multi-objective and noisy optimization procedure \cite{karl_multi-objective_2023}. In the realm of multi-objective optimization, evolutionary algorithms have proven their effectiveness. Among them, NSGA-II \cite{deb_fast_2002} is particularly popular. This genetic algorithm is built upon non-dominated sorting and an elitism criterion based on that. However, elitism makes this algorithm sensitive to noise: if observations are wrongly overestimated, the entire optimization process can be influenced. 

There are three different kinds of noise that can appear in multi-objective optimization \cite{zhou_empirical_2023}. The first kind of noise occurs when decision variables cannot be precisely controlled and is addressed within robust optimization. An unstable environment that changes its behavior over time produces the second type and is handled by methods of dynamic optimization. The third kind of noise results from a stochastic objective function and is referred to as noisy optimization. This type of optimization process occurs, for example, in hyperparameter optimization in machine learning, therefore this paper will focus on this kind of noise. The optimization of noisy objective functions involves resource allocation; reevaluation reduces uncertainty in performance estimates while simultaneously decreasing the budget available for further optimization. There are many strategies to make the NSGA-II algorithm resistant to noise, which can be found in \cite{rakshit_noisy_2017}. The rolling tide EA \cite{fieldsend_rolling_2015} is one of the few evolutionary algorithms specifically designed to handle multiple noisy objective functions.

If hyperparameter tuning aims solely at optimizing one performance measure, such as the mean squared error, established methods like F-Race \cite{birattari_racing_2002} and proposed sequential tests \cite{buczak_using_2024} utilize statistical methods to decide whether to resample or not. For multi-objective optimization, a test procedure based on a normal assumption has been proposed to determine whether resampling is necessary \cite{syberfeldt_evolutionary_2010}. Another way of incorporating the stochasticity of the objective functions is the probability of dominance\cite{teich_pareto-front_2001}, which aims to estimate the probability of each domination relation. \\

In this paper, we propose a new resampling decision function that is built upon the Probability of Domination. Instead of making assumptions about the distributions as done by \citeauthor{teich_pareto-front_2001}, we suggest using bootstrap for distribution estimation. The normal bootstrap procedure requires a certain number of samples, but resampling each poor performing point a second time is very costly. By utilizing the assumption of homoscedasticity in the case of a small sample size, we are able to implement an error-resampling-based bootstrap. The decision whether to resample a point is then made based on the probability of not being dominated by the Pareto front.

The remainder of this paper is structured as follows: The noisy multi-objective optimization problem is defined in the next section along with the probability of dominance, which forms the formal basis for this paper. Subsequently, NSGA-II with different noise handling strategies is presented. In the fourth part, we define our new approach. Therefore, the bootstrap procedure for estimating the distribution of sample means is presented and its application for estimating the probability of dominance is shown. In the subsequent part, a simulation study is presented, which compares the new resampling algorithm with other NSGA-II resampling strategies as well as the rolling tide EA. 

\section{Noisy Multi-Objective Optimization Problems (NMOOP)}
In Multi-Objective Optimization Problems (MOOP) the goal is to find the set of pareto-optimal points. In the following the MOOP is defined analogous to \cite{teich_pareto-front_2001}. A MOOP consists of a set of $D$ decision variables $\mathbf{x} = (x_1,...,x_D)$ and a set of $T$ objective functions \linebreak $\mathbf{g} = (g_1,...,g_T);\; \; g_t: \mathcal{X} \to \mathbb{R}$. For simplicity of the notation we will name the image of the target functions $\mathbf{y} = \mathbf{g}(\mathbf{x})$ and the image space $\mathcal{Y}$. The optimization goal is w.l.o.g.
\[
\min \mathbf{y} = \mathbf{g}(\mathbf{x}) = (g_1(\mathbf{x}),...,g_T(\mathbf{x})) \; .
\]
For any two decision vectors $\mathbf{x}_i, \mathbf{x}_j$ one the following domination relationships holds, which are defined as
\begin{description}
    \item[$\mathbf{x}_i \succeq \mathbf{x}_j$:] $\mathbf{x}_i$ weakly dominates $\mathbf{x}_j$ if $\forall t \in \{1,...,T\}: g_t(\mathbf{x}_i) \le g_t(\mathbf{x}_j)$
    \item[$\mathbf{x}_i \succ \mathbf{x}_j$:] $\mathbf{x}_i$ dominates $\mathbf{x}_j$ if $\forall t \in \{1,...,T\}: g_t(\mathbf{x}_i) \le g_t(\mathbf{x}_j), \; \exists t \in \{1,...,T\}: g_t(\mathbf{x}_i) < g_t(\mathbf{x}_j)$
    \item[$\mathbf{x}_i \sim \mathbf{x}_j$:] $\mathbf{x}_i$ is indifferent to $\mathbf{x}_j$ iff $\mathbf{x}_i \nsucc \mathbf{x}_j$ and $\mathbf{x}_j \nsucc \mathbf{x}_i$.
\end{description} 

If a decision vector \(\mathbf{x}_i \in \mathcal{X}\) is not dominated with respect to a set \(S \subseteq \mathcal{X}\), it is called Pareto-optimal.  That means, that there does not exist any \(\mathbf{x}_j \in S\) such that \(\mathbf{x}_j \succ \mathbf{x}_i\). All decision points that are not dominated with respect to the entire decision space form the Pareto-optimal set.

In the case of a noisy multi-objective optimization problem (NMOOP), it is assumed that the objective function \(\mathbf{g}\) is of stochastic nature. Therefore, evaluating the objective function at a certain point in the objective space \(\mathcal{X}\) results in a random variable \(\mathbf{Y} = \mathbf{g}(\mathbf{x}) = \mathbf{\mu}(\mathbf{x}) + \mathbf{e}\). In this notation, \(\mathbf{\mu}(\mathbf{x})\) represents the mean function we want to optimize, while \(\mathbf{e}\) denotes the error terms, with expected value of $0$.
The distribution of a particular \(\mathbf{Y}_i\) is given by an unknown density function \(f(\mathbf{Y}_i)\). In this paper, we focus on NMOOPs (Noisy Multi-Objective Optimization Problems) that aim to find the Pareto front of the expected values
\[
\underset{\mathbf{x} \in \mathcal{X}}{\min} \; E[\mathbf{g}(\mathbf{x})] = \underset{\mathbf{x} \in \mathcal{X}}{\min} \; \mu(\mathbf{x}) \; .
\]

If the objective functions are stochastic, evaluating a certain set of decision variables is a sampling process of random variables. If we compare the evaluation of to decision points, there may be a probability for each of the dominance relations, so that holds  
\[
P(\mathbf{x}_i \succ \mathbf{x}_j) + P(\mathbf{x}_i \prec \mathbf{x}_j) + P(\mathbf{x}_i \sim \mathbf{x}_j) = 1 \; .
\]

As the evaluation of the target functions just returns a random variable, we can only roughly estimate the expected value at the point of evaluation. For improving the precision, a point can be evaluated multiple times, these multiple realizations will be called $\mathbf{y}^{(1)},...,\mathbf{y}^{(N)}$ and its mean, which is used as an estimate of the expected value $\mathbf{\bar{y}}$.

\section{Resampling Strategies for NMOOP}
The non-dominated sorting genetic algorithm, known as NSGA-II \cite{deb_fast_2002}, is one of the most widely used evolutionary algorithms for multi-objective optimization problems. It is a population-based genetic algorithm with elitism. In the NSGA-II algorithm, the current population is used to create an offspring population of the same size. Each offspring is created through mutation and crossover operators applied to two parents, which are selected via tournament selection. The tournament criterion is first the Pareto rank and second the density of solutions in the objective space around the individual. After offspring creation, the combined population is sorted using the same criteria as in tournament selection, and the worse half is discarded. The strength of this algorithm lies in its ability to preserve variation in the gene pool while stochastically selecting better candidates more frequently for offspring creation.

Each classical multi-objective optimization algorithm is affected by noise to some extent, as objective estimates can be inaccurate. Optimization algorithms that rely on elitism are particularly influenced during the optimization process. The overestimation of a point can lead to an incorrect Pareto rank, which subsequently affects offspring selection due to elitism. Without resampling, overestimated points may remain permanently in the Pareto front, thereby hindering the optimization process. 

There are several resampling approaches to make NSGA-II robust against noise, which can be classified into different categories. One-shot resampling methods assign a point a fixed number of resamples that will not change, while sequential resampling strategies periodically decide whether a point should receive additional resampling budget. The decision regarding the budget can be based on stochasticity, specifically the variance of a point, or it can be independent of this property (see \cite{rakshit_noisy_2017}). Important algorithm-independent sampling functions include \textit{static resampling}, \textit{time-based resampling} \cite{siegmund_comparative_2013}, \textit{domination-strength resampling} \cite{zitzler_multiobjective_1999} and \cite{siegmund_comparative_2013}, \textit{rank-based resampling} \cite{siegmund_comparative_2013}, and \textit{standard error dynamic resampling} \cite{di_pietro_applying_2004}. \\

\subsection{Static Resampling}
The simplest idea of handling noise is, to evaluate each point multiple times. This approach is known under the name \textit{static resampling} and only has one parameter, the number of evaluations for each point $N$. The normal NSGA-II without resampling can be seen as one version of it, with a budget of $N = 1$ evaluations for each point. The downside of this approach is obviously that points that seem bad are evaluated multiple times if $N > 1$. 

\subsection{Dynamic Resampling}
The dynamic resampling approaches try to overcome the inefficiency of static resampling, by considering the properties of the decision point. The most considered properties are the variance at the decision point, the domination relation to the other points, or the progress in the optimization process. The decision of whether or not to resample a point $\mathbf{x_i}$ is then made with a decision function $\Delta(i)$. In the following we present multiple resampling decision function, which will return $T$ for resampling the $i$th point and $F$ if not.

Some of the advanced approaches are defined by \citeauthor{siegmund_comparative_2013} \cite{siegmund_comparative_2013} based on a fraction $\nu$ of a predefined maximal number of resampling, resulting in the following decision function
\[
\Delta(i) = \begin{cases}
    TRUE \quad \text{while} \quad n_i < \nu_i \cdot N \\
    FALSE \quad \text{else} \quad . 
\end{cases}
\]
The individual fraction of the maximal budget $\nu_i$ can be determined by different approaches.

\textbf{Domination-strength resampling} aims to evaluate those points multiple times that dominate many other points. The strength metric \cite{zitzler_multiobjective_1999} summarizes the fraction of observations dominated as follows:
\[
\text{strength}(x_i) = \frac{1}{|P|} \sum_{x_j \in P} \mathbbm{1}_{x_i \succeq x_j} 
\]
where \(P\) denotes the population, and \(|P|\) is its size. The domination-strength resampling approach \cite{siegmund_comparative_2013} is based on this strength metric and is defined by the following formulas:
\[
\nu_i = 
\begin{cases}
    \frac{\text{strength}(x_i)}{\underset{j}{\max} ( \text{strength}(x_j))} & \text{for} \; \underset{x_j \in P}{\max}  (\text{strength}(x_j)) > \frac{1}{|P|} \\
    \max(0, \frac{\text{strength}(x_i)}{\underset{x_j \ in P}{\max} ( \text{strength}(x_j))}) & \text{for} \; \underset{x_j \in P}{\max}(\text{strength}(x_j)) = \frac{1}{|P|}\\
    1 & \text{for} \; \underset{x_j \in P}{\max}  (\text{strength}(x_j)) = 0 \; .
\end{cases}
\]
This strategy will resample a point $x_i$ the maximal number of times $N$, if $x_i$ is of maximal strength or if all points have strength 0.\\ 

The \textbf{rank-based resampling} approach also aims to allocate more budget to the observations in the top ranks while linearly distributing the budget among the ranks. This is achieved by using the following formula
\[
\nu_i = 1 - \frac{r(x_i) - 1}{R - 1} 
\]
where \(R\) represents the maximum pareto-rank among the population and $r(x_i)$ the Pareto-rank of $x_i$. 

The idea of \textbf{time-based resampling} is that more evaluations are needed as the optimization process approaches the Pareto front. The simple linear time-based budget function is given by

\[
\nu_i = \frac{n_{\text{gen}}}{N_{\text{gen}}}
\]

where \(n_{\text{gen}}\) denotes the current generation number and \(N_{\text{gen}}\) is the maximum number of generations that we are willing to create.\\

The \textbf{standard error dynamic resampling} \cite{di_pietro_applying_2004} aims to resample each point until the standard error of the mean estimate falls below a certain threshold. 
As this approach was originally designed for univariate objective functions, its application in the multi-objective case requires some rethinking. Based on a certain aggregation function \(\text{AGF}\), we can create a univariate error term defined as follows:

\[
se_i = \text{AGF} \left( \sqrt{\frac{1}{N_i - 1} \sum_{n=1}^{N_i}(y_{t}^{(n)} - \bar{y_{t}})^2 } \right) \; .
\]

The decision is then made by

\[
\Delta_{SE}(i) = 
\begin{cases}
    TRUE & \text{if} \quad se_i > se_{\text{thr}} \\
    FALSE & \text{else} \; .
\end{cases}
\]
The aggregation function might be either a maximum or a (weighted) mean.

\subsection{One-Shot and Sequential Resampling}
Besides the decision of how much budget a point receives, the second question is whether this budget is allocated all at once—\textit{one-shot resampling}—or piecewise—\textit{sequential resampling}. One-shot resampling means that a point is evaluated a certain number of times during the iteration in which it is created, based on the decision functions above, and is never checked again for additional budget allocation.

The sequential resampling strategies for NSGA-II involve iterative checks to determine whether a specific decision point needs to be resampled; therefore, in every generation, each point is checked to see whether it meets the criteria of the decision functions. This procedure can be described using the pseudocode in \ref{alg:seqNSDGA2}. The one-shot algorithm is quite similar, but instead of checking the entire population, only the offspring are considered, and they are evaluated as long as the decision function dictates.

Each of the resampling strategies presented above can be applied in either the one-shot or the sequential resampling approach. In the simulation, we apply the dynamic decision functions within the sequential resampling scheme because it tends to be more efficient. The static approach is applied in the one-shot resampling strategy.

\begin{algorithm}[hb]
    \caption{Sequential Resampling NSGA-II}\label{alg:seqNSDGA2}
    \begin{algorithmic}
    \REQUIRE popSize, nGen
    \STATE initialize\_population
    \FOR{i in 1:nGen}
        \STATE createOffspring
        \STATE evaluateOffspring
        \FORALL{p in pop} 
            \IF{$\Delta(\text{p})$}
                \STATE evaluate(p)
            \ENDIF
        \ENDFOR
        \STATE selectSurviver
    \ENDFOR
    \STATE pareto\_front(pop)
    \end{algorithmic}
\end{algorithm}

\subsection{Rolling Tide Evolutionary Algorithm}
The Rolling Tide Evolutionary Algorithm (RTEA) \cite{fieldsend_rolling_2015} is a multi-objective optimization algorithm designed for noisy objective functions. RTEA is built upon strong elitism and repeatedly resamples the points identified as elite. The RTEA requires parameters for the total budget \( m \), the number of resamples per iteration \( k \), the size of the initial sample \( p \), and the proportion of evaluations allocated for refinement \( z \). The RTEA consists of three parts: initialization, optimization, and refinement. Without implementation details, the algorithm can be described as follows:

For the \textbf{initialization}, \( p \) points in the decision space are sampled and evaluated once. An initial Pareto front \( F \) is extracted, while the remaining points are stored in an archive. 

The \textbf{optimization} phase consists of the alternating creation of an offspring and the re-evaluation of a portion of the Pareto front. For the creation of the offspring, two members of the Pareto front are selected. A child is generated based on these parents using crossover and mutation. The offspring is then evaluated once. Afterwards, the Pareto front and archive are updated. Subsequently, \( k \) points from the current estimate of the Pareto front are resampled and evaluated. The Pareto front and archive are updated again. The optimization process ends when \( (1 - z) \cdot m \) evaluations have been performed.

The final \( z \cdot m \) evaluations of the budget are used for the \textbf{refinement} of the Pareto front.

    
    

\section{Bootstrap Based Testing}
The rank- and domination-based resampling strategies presented above are based on the idea that it is important to be more certain about the points that are likely to remain in the population and be chosen as parents for the next generation. The downside is that these approaches do not account for the variability of the evaluations. Standard error-based resampling, on the other hand, considers only the variability and not the rank of the points.

In this section, we will derive a resampling strategy that considers both the position in the overall rank and the variability in the estimate. A natural way to merge the ideas of considering both rank and variability is offered by the probability of domination.

The theoretical considerations of the probability of being dominated \cite{teich_pareto-front_2001} are derived under the assumption that the evaluations of the objective functions at a given point are independent and uniformly distributed. Instead of relying on such statistical assumptions, it is possible to use statistical bootstrap methods to estimate the distribution of a sample statistic. These distributions can be used to estimate the probability that a certain point will dominate a member of the Pareto front, which serves as our metric to determine whether resampling a certain point is worthwhile. As we bootstrap the sample mean and compare these distributions, the variability represented in the distributions decreases as more evaluations are conducted.

\subsection{The Bootstrap Distribution of the Sample Mean}
Bootstrapping is a statistical method, which is mostly used to estimate distributions of test statistics or parameter estimates. For estimating the distribution of the sample mean, we assume \(\mathbf{Y}^{(1)}, \ldots, \mathbf{Y}^{(N)}\) to be independent and identically distributed \(D\)-dimensional random variables. The mean of these variables, denoted as \(\mathbf{\bar{Y}}\), is itself a random variable and its distribution can be estimated using bootstrapping. To obtain \(B\) bootstrap estimates of the mean, an artificial sample of size \(N\) is drawn with replacement from the real observations:

\[
\mathbf{\tilde{\bar{Y}}}^b = \frac{1}{N} \sum_{n=1}^N \mathbf{\tilde{Y}}^{(n)} \quad \text{with} \quad \mathbf{\tilde{Y}}^{(n)} \sim \hat{F}_{y}
\]

where \(\hat{F}_{y}\) is the empirical distribution of \(\mathbf{y}_1, \ldots, \mathbf{y}_N\) (see \cite{dikta_bootstrap_2021}[p. 22 seqq.]). The bootstrap distribution is known to underestimate the variance by a factor of \(\frac{N}{N-1}\). Adjusting for this underestimation of the variance, the sampling scheme can be rewritten as follows:

\[
\mathbf{\tilde{\bar{Y}}}^b = \frac{1}{N} \sum_{n=1}^N (\mathbf{\bar{y}} + \sqrt{\frac{N}{N-1}}\, \mathbf{\tilde{E}}^{(n)})  \;  \text{with} \; 
\mathbf{\tilde{E}}^{(n)} \sim \hat{F}_{e}
\]

and \(\hat{F}_{e}\) being the empirical distribution of the dispersion terms \(\mathbf{e}^{(i)} = \mathbf{y}^{(i)} - \mathbf{\bar{y}}\).
The empirical distribution of the \(B\)-bootstrap samples will be referred to as \(\hat{F}_{\mathbf{\tilde{\bar{Y}}}}\) and its corresponding probability density function as \(\hat{f}_{\mathbf{\tilde{\bar{Y}}}}\). These distributions will serve as the basis for estimating the probability of domination.

\subsection{Estimating the Probability of Dominating}
For two random variables $Y_i, Y_j$ with a joint density function $f_{Y_i, Y_j}$ the probability of domination is generally given by 
\[
P(x_i \succ x_j) = \int_{\mathcal{Y}} \int_{\mathcal{Y}} f_{Y_i, Y_j} (y, z) \mathbbm{1}_{y_d < z_d \forall d}  \; \text{d}y  \; \text{d} z \; .
\]
This is nothing other than the probability mass falling into the part of $\mathcal{Y} \times \mathcal{Y}$ in which $x_i \succ x_j$ holds. If we assume that these random variables are independent, then the joint density is the product of the individual densities. This results in
\[
P(x_i \succ x_j) =  \int_{\mathcal{Y}} \int_{\mathcal{Y}} f_i(y) f_j(z)  \mathbbm{1}_{y_d < z_d \forall d}   \; \text{d}y  \; \text{d} z \; .
\]
If we do not know the distributions of these random variables but have samples of both, we can estimate the probability of domination using the proportion of observed instances where this domination relation holds
\[
\hat{P}(x_i \succ x_j) = \frac{1}{N_i N_j} \sum_{n_i}^{N_i} \sum_{n_j}^{N_j} \mathbbm{1}_{y_{i,d}^{(n_i)} < y_{j,d}^{(n_j)} \; \forall d} \; .
\]
In our case, we want to estimate the probability that the sample mean of one decision point dominates that of another. With bootstrap, we can obtain samples of these distributions and use them to estimate the probability of domination using the form above.

\subsection{Limitation of the Bootstrap}
The bootstrap distribution of the mean estimate has limitations, primarily due to its statistical properties and the requirement for a certain amount of data. The number of different bootstrap samples is limited by the number of observations we have for that decision point: In the case of only one observation, there is only one bootstrap sample to draw; in the case of two observations, there are three combinations to draw without considering the order; and in the case of three observations, there are ten samples. In general, the number of different bootstrap samples is given by the formula $\binom{2N - 1}{N}$.

In the optimization process, we will mostly have $N \ll 5$, and for efficiency, it is desirable not to evaluate poor points even a second time. Therefore, we will make the assumption that points we have not observed frequently behave similarly to the points we already know. Under the naive assumption of homoscedasticity across the entire space, we can define a set of dispersion terms as follows:
\[
\mathcal{E} = \bigcup_{i=1}^I \bigcup_{n=1}^{N_i}  \sqrt{\frac{N_i}{N_i-1}}(\mathbf{y}_{i}^{(n)} - \mathbf{\bar{y}}_{i}) .
\]

Assuming that the evaluated decision points only differ in their expected values, resampling from $\mathcal{E}$ is stochastically equivalent to the bootstrap procedure described above. If we have only one observation, it seems natural to assume that the point just behaves like all the other points we have seen. For a growing number of observations, it is desirable to weight the variability in this point more and more over the variability in other points. A strategy which aims for this property is that we draw for each bootstrap sample a certain number of observations for the mean calculations from the over all dispersion set $\mathcal{E}$ and the rest from the dispersion of the point we are bootstrapping. So the simple strategy we will employ is that we draw 1 summand of each bootstrap mean sample from the over all dispersion and $N-1$ summands from the seen variability of the point.

\[
\mathbf{\tilde{\bar{Y}}}^b = \mathbf{\bar{y}} + \frac{1}{N}\left(  \mathbf{\tilde{E}} + \sum_{n=1}^{N-1} \sqrt{\frac{N}{N-1}} (\mathbf{\tilde{Y}}^{(n)} - \mathbf{\bar{y}})\right) 
\]

with 

\[
\mathbf{\tilde{E}} \sim F_{\mathcal{E}}, \quad \mathbf{\tilde{Y}}^{(n)} \sim F_{y}.
\]
In cases where a random variable has been evaluated only once, we rely solely on the global dispersion set. However, as the number of evaluations increases, individual variation increasingly dominates.

\subsection{Adaptive Resampling with Bootstrap}
Let $\hat{S}$ be our current estimate of the Pareto front each with an individual number of evaluations. A point which is not in the Pareto front can be seen as promising, if the probability, that the sample mean of this point dominates any point of the Pareto front with at least a certain probability.  Therefore, we can define a function for deciding, whether to resample the point \(x_j\) with
\[
\Delta_{PD}(\hat{S}, j) =  
\begin{cases} 
FALSE & \quad \text{if} \quad \underset{\mathbf{x}_s \in \hat{S}}{\max} \hat{P}(\mathbf{x}_j \succeq \mathbf{x}_s) > \alpha_u \\ 
FALSE & \quad \text{if} \quad  \underset{\mathbf{x}_s \in \hat{S}}{\max} \hat{P}(\mathbf{x}_j \succeq \mathbf{x}_s) <  \alpha_l \\ 
TRUE &  \quad  \text{else} \; .
\end{cases}
\]
The lower bound \(\alpha_l \in (0, 0.5)\) defines a threshold for the probability that a point will dominate a member of the Pareto front. It represents the minimum potential that a point must exhibit. In contrast, the parameter \(\alpha_u \in (0.5, 1]\) represents an upper threshold, where we assume that further resampling would be a waste of resources due to high confidence in current results. If in the following we need a short term for this resampling strategy, we will call it \textit{ARB}.

\subsection{Algorithmic Considerations}
The estimation of the probability of domination as defined above is done with bootstrap samples of the sample mean. As we only need a rough estimate of the probability, we will stick to 100 bootstrap samples. To be able to draw also enough different samples in the case of having evaluated a certain point only once, it is necessary to ensure that the set of errors $\mathcal{E}$ matches at least that size. We will force this set to be 100 in size, so we could draw 100 different samples from it. In the initialization phase, therefore, a population of size $M_i > 100$ is initialized and the 'best' 100 in the sense of NSGA-II are evaluated a second time. The differences between the second evaluation and the mean of both scaled with the factor $\sqrt{\frac{N}{N-1}}$ is the starting error base. While the algorithm is running, we will ensure that the error base consists of the 100 last errors observed, so that we estimate the dispersion at a point based on the last re-evaluated points before, which will most likely of the generation of its parents and share some similarities. It must also to be ensured that the mean of the error base is 0 before sampling. 

\section{Simulation Setting and Metrices}
For the evaluation of the developed method, we conducted a simulation study. In this simulation study, we compare our resampling strategy with the NSGA-II resampling strategies presented, as well as with the RTEA. For a meaningful comparison, we consider two types of noise: Gaussian and $\chi^2$. While Gaussian noise is often assumed in many scenarios, this assumption may often not hold, especially if we consider a setting like hyperparameter optimization, where the error size is bounded on one side. For the case of one-sided bounded and skewed distributions, we also investigated $\chi^2$-distributed noise.

\subsection{Data Generation}
As a basis for the comparison, we use the UPC functions \cite{zhang_multiobjective_2009}, which serve as the mean function with noise added to each evaluation. The data generation can be described as

\[
Y_{d} = g_d(\mathbf{x}) + v_d(\mathbf{x}) \varepsilon  
\]

with either $\varepsilon \sim N(0,1)$ or $\varepsilon \sim \frac{1}{\sqrt{2k}}(\chi^2_k - k)$ with $k \in \{1,2\}$. The function $v_d(\mathbf{x})$ scales the noise up and down for the heteroscedastic case; in the homoscedastic case, it is simply a constant. As the error distributions are transformed in such a way that the standard deviation is equal to one, we choose $v_d$ to be constant with the desired standard deviation of $\sigma \in \{0.01, 0.1, 0.5, 1, \sqrt{2}, 2\}$ for both types of $\chi^2$-noise as well as for Gaussian noise. In addition, we perform a run with no noise.

\subsection{Quality Measures}
The goal of multi-objective optimization is to find points as near as possible to the true Pareto front. For measuring coverage, uniformity, and spread, the inverse generational distance (IGD) and the dominated hypervolume are well suited \cite{laszczyk_survey_2019}. An optimization algorithm will return a set of points $S$, which to him seem not to be dominated. Based on knowledge of the true mean function $\mathbf{\mu}(\mathbf{x})$, we can generate a set $\hat{S}$, which consists of the decision points, which expected values are not dominated
\[
\hat{S} = \{\mathbf{x}_i \in S:  \mathbf{\mu}(\mathbf{x}_i) \succeq \mathbf{\mu}(\mathbf{x}_j)  \wedge \mathbf{\mu}(\mathbf{x}_i) \sim \mathbf{\mu}(\mathbf{x}_j) \: \forall x_j\in S \} \; .
\]
Based on this set we can estimate the performance of a solution.

The \textbf{HyperVolume} \cite{zitzler_multiobjective_1999} measures the volume between a nadir point \(u^{nadir}\) and the estimated Pareto set \(\hat{S}\). Based on the Lebesgue measure \(\Lambda\), the raw HV is defined as

\[
\text{HV}_{\text{raw}}(\hat{S}) = \Lambda\left( \bigcup_{ \mathbf{s} \in \hat{S}}\{\mathbf{s'}| \mathbf{s} \prec \mathbf{s'} \prec u^{nadir} \}  \right).
\]

The HV can be seen as a measure of improvement over the nadir point. For the purpose of comparability, we scale the \(\text{HV}_{\text{raw}}\) by the hypervolume of the true Pareto front:

\[
\text{HV} = \frac{\text{HV}_{\text{raw}}(\hat{S})}{\text{HV}_{\text{raw}}(PF)}.
\]

While the HV serves as a measure of improvement, it weights different directions of progress quite unequally. The most important angle of improvement for the HV is directly in the direction of the \(0\)-point, while finding optimal points only leads to marginal improvements in the HV.

The \textbf{Inverted Generational Distance} \cite{bosman_balance_2003}, on the other hand, measures how well the full length of the Pareto front is covered by a certain set.  The power mean variation of the inverse generational distance is defined as 

\[
\text{IGD}_p(\hat{S}, \text{PF}) = \left( \frac{1}{|\text{PF}|} \sum_{\mathbf{u} \in \text{PF}} \text{dist}(\mathbf{u}, \hat{S})^p \right)^{1/p}
\]

with \(\text{PF}\) being a sample from the true Pareto front. The \textit{IGD} measures how well the entire Pareto front is approximated by the optimization algorithm.

\subsection{Parameterization}
For comparison of the different resampling approaches, we run them with different parameter settings.
The max samples parameter of the NSGA-II resampling algorithms is varied in $\{5, 10, 20\}$ and for the standard error based resampling procedure we used  $\sigma_{thr} \in \{0.01, 0.05, 0.1\}$. All NSGA-II variants were restricted to a popsize of 40. As we also aimed to investigate the behavior of our new proposed resampling scheme, we also varied the upper and lower thresholds in a range of $\alpha_l \in \{0.1,0.2,0.5\}$ and $\alpha_u \in \{0.75, 0.9, 1.0\}$. For the purpose of comparability, the mutation and crossover operations are the same for all competing algorithms. For the static resampling approach, we used $N \in \{1, 5, 10, 20\}$. For the RTEA the number of resamples $k$ per iteration is set to one. The starting population size is $40$ and the refinement fraction $z$ is set to $0.1$. 

\section{Evaluation}
In this section, we present the simulation results. As we ran different NSGA-II variations, including our bootstrap-based resampling algorithm, each with multiple parameter settings, careful consideration must be given to comparisons. Comparing only the best-performing algorithm in each family would inadvertently introduce selection bias. 

To avoid this, we compare the results in two unbiased ways: In the first comparison, we randomly split the 30 replications into two parts—one for selecting the best parameter setting and the other for comparison purposes. To account for potential effects of this random split, we repeat this procedure multiple times. Since this scenario is highly unrealistic, we conduct a second comparison in which an initial small run of 5,000 evaluations is used to select the best-performing parameter combination. These results are then compared with those from a full run comprising 50,000 evaluations.

In our evaluation, we group all resampling strategies together to facilitate comparison between RTEA, NSGA-II with a static resampling scheme, NSGA-II with one of the dynamic strategies mentioned above, and our proposed strategy. All dynamic resampling strategies presented above were grouped together, as their overall performance was mediocre. The chosen resampling strategy was the one that performed best in the parameter selection step.

All comparisons are based on the HV metric, as comparisons using IGD did not yield substantially different results.

\subsection{Comparing with Optimal Parameters}
\begin{figure}
    \centering
    \includegraphics[width=0.75\linewidth]{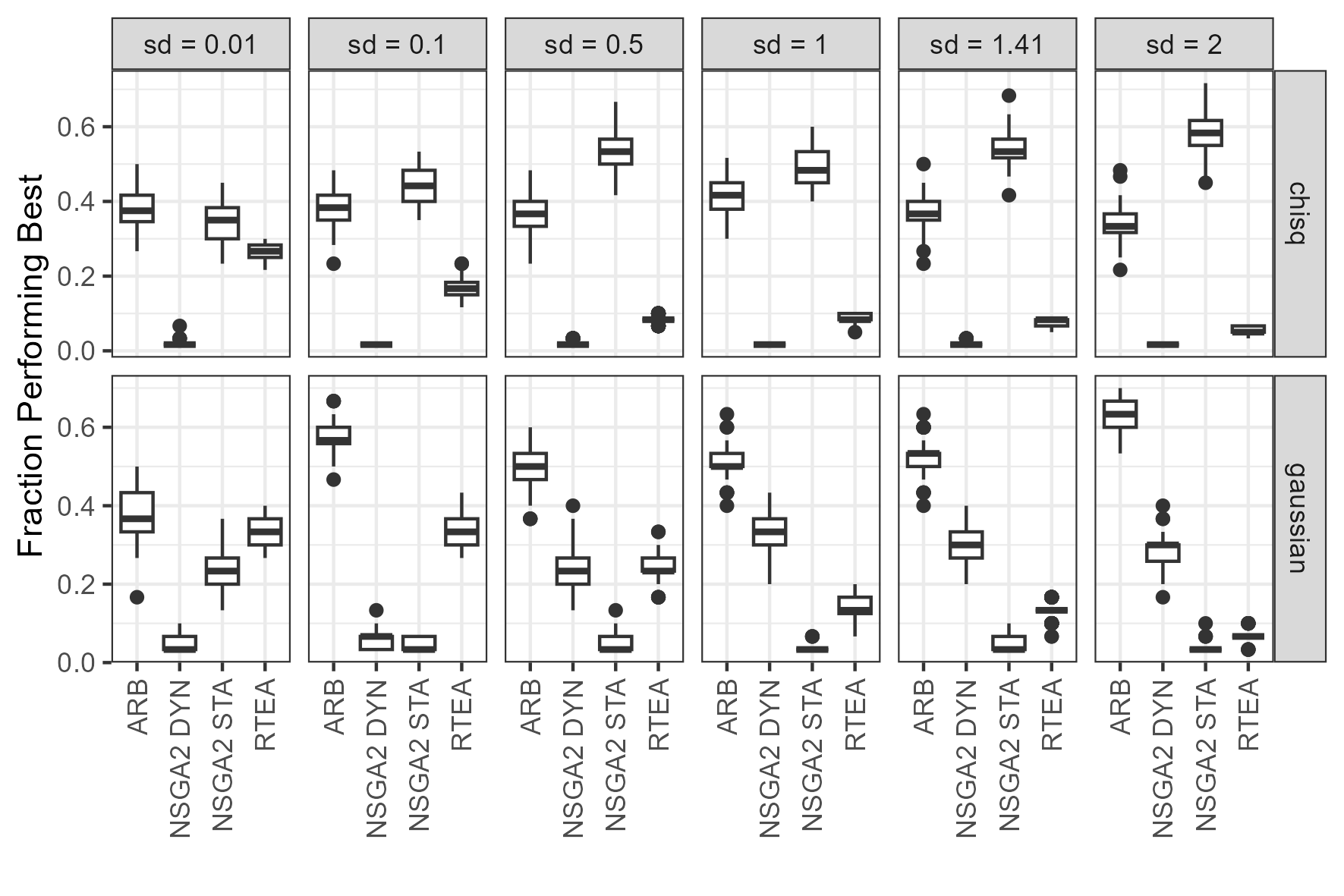}
    \caption{Fraction of performing best in terms of dominated hypervolume of the different clusters of resampling strategies by using optimized parameters for each setting. The boxplots are representing the distribution of the fraction based on 100 random splits between selection replications and comparison replications. For the sake }
    \label{fig:Fraction_Best_IDX_VAR_SD}
\end{figure}

One approach to comparing the algorithms without bias is to use 20 replications to select the best parameters for a problem setting and the last 10 replications to compare their performance. We repeat this procedure of random splitting 100 times and then investigate how often each algorithm performs best in a simulation setting. The results are shown in Figure \ref{fig:Fraction_Best_IDX_VAR_SD}. In all settings with Gaussian and $\chi^2$ noise, across all standard deviations, the new resampling approach outperforms both the RTEA and the dynamic NSGA-II in many cases. In the settings with $\chi^2$ noise, the static resampling NSGA-II, which never resamples, performs best in most situations. The reason for this behavior lies in the distribution of the noise. Since the noise is restricted in the direction of interest, it is not possible to significantly overestimate the performance of a single point. With $\chi^2$ noise, overestimation does not affect performance as severely because it is quite likely that other points are similarly overestimated. 

Even after grouping all the dynamic resampling approaches together ad selecting the best of them, the whole group of strategies performed poorly, as can be seen in Figure \ref{fig:Fraction_Best_IDX_VAR_SD}. Only in the case of Gaussian noise with a relatively high variance do these strategies begin to win in some portion of the settings.

\subsection{Comparing with Pre-Study}
As the comparison above involves a full-size parameter optimization step beforehand, it does not represent a realistic scenario. A more realistic approach would be to use 5,000 evaluations to determine which parameters an algorithm should use. The number of settings and replications in which each resampling strategy performs best is shown in Table \ref{tbl:win_5000}.

\begin{figure}[ht]
    \centering
    \includegraphics[width=0.75\linewidth]{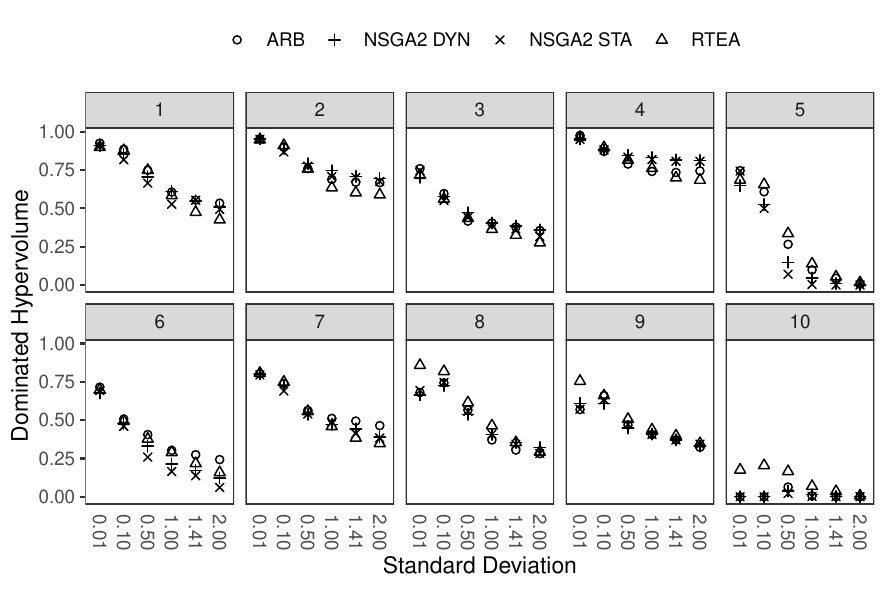}
    \caption{Average dominated Hypervolume over 30 replication with 50.000 evaluations based on a parameter optimization step with 5000 evaluations in the setting with Gaussian noise.}
    \label{fig:HV_NV}
\end{figure}

\begin{figure}[ht]
    \centering
    \includegraphics[width=0.75\linewidth]{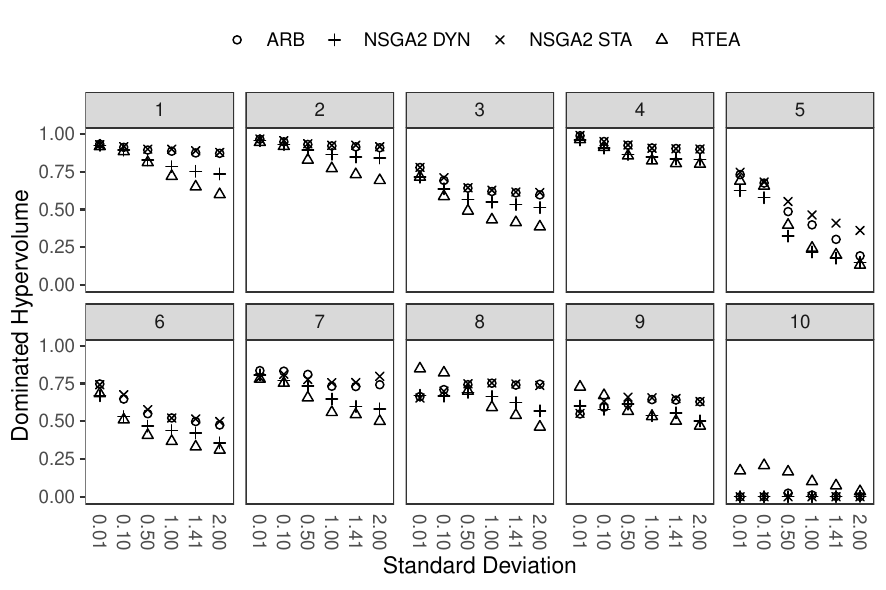}
    \caption{Average dominated Hypervolume over 30 replication with 50.000 evaluations based on a parameter optimization step with 5000 evaluations in the setting with $\chi^2$ noise with df = 1.}
   \label{fig:HV_Chisq1}
\end{figure}

\begin{figure}[ht]
    \centering
    \includegraphics[width=0.75\linewidth]{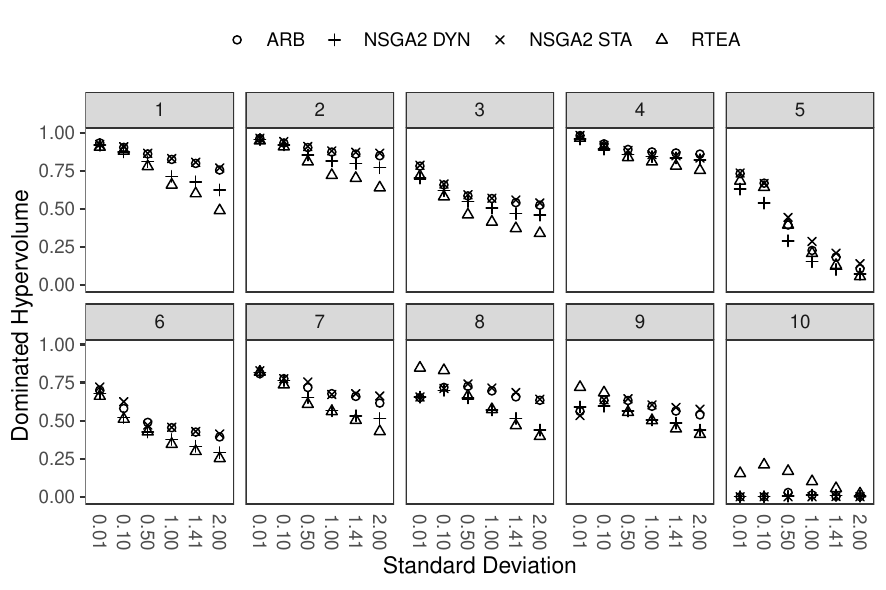}
    \caption{Average dominated Hypervolume over 30 replication with 50.000 evaluations based on a parameter optimization step with 5000 evaluations in the setting with  $\chi^2$ noise with df = 2.}
   \label{fig:HV_Chisq2}
\end{figure}

\begin{table}[ht] \caption{Counts of settings in which the resampling strategy performs best after 50,000 evaluations in terms of dominated hypervolume, following the use of the first 5,000 evaluations for estimating the performance of different algorithm parameters.} 
\centering
\begin{tabular}{rrrr}
  \hline
 & $\chi^2$ & Gaussian & none \\ 
  \hline
ARB & 1194 & 549 & 107 \\ 
  NSGA-II DYN & 175 & 542 &  17 \\ 
  NSGA-II STA & 1614 & 280 & 104 \\ 
  RTEA & 617 & 609 &  72 \\ 
   \hline
\end{tabular}
\label{tbl:win_5000}
\end{table}
The results demonstrate the high flexibility of the proposed resampling strategy, as it ranks second in the cases of Gaussian noise, $\chi^2$ noise, and even in the absence of noise. While the static resampling approach performs best in the case of $\chi^2$ noise or no noise, it performs very poorly in the case of Gaussian noise, whereas the RTEA exhibits the opposite behavior. Figures \ref{fig:HV_Chisq1}, \ref{fig:HV_Chisq2}, and \ref{fig:HV_NV} show the average performance across different test functions and standard deviations. The performance differences between the fixed resampling strategy and our proposed resampling strategy, on one hand, and the dynamic strategy or RTEA, on the other hand, are particularly striking in the case of $\chi^2$ noise.

The primary reason for this likely lies in the fact that the latter strategies allocate a fraction of their total budget to resampling, while the fixed resampling strategy does not; our proposed strategy adapts to the situation. In the case of Gaussian noise, the RTEA demonstrates its strength with its strict elitism and focus on correctly estimating members of the Pareto front.

By using simulations with 5,000 evaluations to select the best parameter settings, we can use 30 replications of 50,000 evaluations to perform statistical tests on differences between resampling strategies without introducing bias. Testing across all simulated scenarios led to results indicating that our proposed resampling strategy performs significantly better than RTEA in approximately $64.7\%$ of cases while being significantly worse in $24.7\%$. The other sequential resampling strategies were outperformed in $71.0\%$ of cases while performing better in about $8.7\%$. The one-shot resampling strategy, which samples a fixed number of points, is the only strategy that outperforms our proposed approach across all simulated settings, with $22.4\%$ versus $20.7\%$. While RTEA tends to perform better than our strategy in scenarios involving Gaussian noise, static resampling performs better when $\chi^2$ noise is present. These fractions should be interpreted with caution as they depend on specific simulated settings.

\subsection{Overall Best Parameters}
An even more desirable property of an optimization algorithm would be to perform well out of the box. Therefore, we use 20 replications to find the overall best parameter setting and compare it with the other resampling strategies. Figure \ref{fig:Fraction_Best_OverAll} in the appendix shows that performance, in comparison to the other resampling strategies, highly depends on the algorithm parameters.

\subsection{Effect of Different Noise Types}
In our simulation study, we used Gaussian and $\chi^2$-distributed noise with the same standard deviations. Comparing the performance of the algorithms on these slightly different problems reveals a few interesting findings:

First, all algorithms performed better with $\chi^2$ noise, despite the standard deviation of the noise being the same in both settings. Since the distributions used (with 1 or 2 degrees of freedom) are bounded on the lower side, an evaluation cannot be highly overestimated.

Since both NSGA-II and RTEA rely on elitism, being too optimistic about the performance of a point causes more harm than being too pessimistic. A point that is incorrectly ranked among the best but actually belongs to the worst in a population will produce new offspring, propagating the error to future generations and thereby affecting convergence speed. The second finding is that our decision procedure can better leverage this effect, whereas RTEA, on the other hand, reevaluates members of the first front even when there is little to gain.

\section{Conclusion}
Noisy multi-objective optimization problems are defined by the constant trade-off between resampling and further exploration. In this paper, we presented a statistical approach based on bootstrapping to estimate the distributions of the sample means of different decision points in the objective space. This approach allows for a statistical comparison between a decision point and the current Pareto front based on the probability of domination. Thus, resampling is restricted to cases where a point seems promising.

The simulation study shows that different noise distributions favor different resampling strategies. In our study, the simple strategy of never resampling and RTEA are the two most exploitative strategies: RTEA performs well with Gaussian noise, while the never-resample strategy excels in cases of bounded noise. The adaptive resampling with bootstraps strategy, however, is flexible enough to perform reasonably well across all settings.

An interesting side finding is the influence of the noise distribution: In the case of $\chi^2$ noise, the best approach is not to resample, even when the standard deviation is large compared with the differences in the mean function. In contrast, with Gaussian noise, even a small standard deviation makes resampling important. The reason for this behavior lies in the shape of the distributions: With $\chi^2$ noise, a point can only be overestimated by a certain amount, and overestimating points is quite probable. An optimistic estimate at a certain point will likely be dominated by the evaluated performance of a sample point with a better expected value since we will also obtain an optimistic sample of that point if we draw multiple times. Gaussian noise, on the other hand, is unbounded, making it possible to obtain highly overestimated points. When drawing many thousands of times, the probability of a better point dominating a highly optimistic estimate with another highly optimistic estimate is quite low. Thus, the noise distribution becomes even more important than merely considering standard deviations.

Although noise is often assumed to be Gaussian-distributed in optimization settings, a skewed noise distribution may often be more accurate. For example, in machine learning, if we aim to optimize an error rate and estimate performance using different train-test splits: some splits may lead to very large errors because an entire region of the problem was assigned to the test split. 

So far, our investigation has only covered homoscedastic noise. Further simulation studies are required to examine how the adaptive resampling with bootstraps performs with heteroscedastic noise.

The strong performance of the ARB on noisy multi-objective optimization problems makes it appealing to transfer this resampling scheme to univariate cases and compare it with F-Race and Successive Halving procedures. Since our proposed procedure does not require a second evaluation for a significant portion of points in NMOOP (Noisy Multi-Objective Optimization Problems), similar benefits should hold for many single-objective problems as well. Therefore, an improvement in performance is expected.

\section*{Funding}
This project was funded by the Ministry of Culture and Science of the State of North Rhine-Westphalia as part of the AI Starter Program.


\printbibliography 

\newpage
\section*{Appendix}

\begin{figure}[ht]
    \centering
    \includegraphics[width=0.75\linewidth]{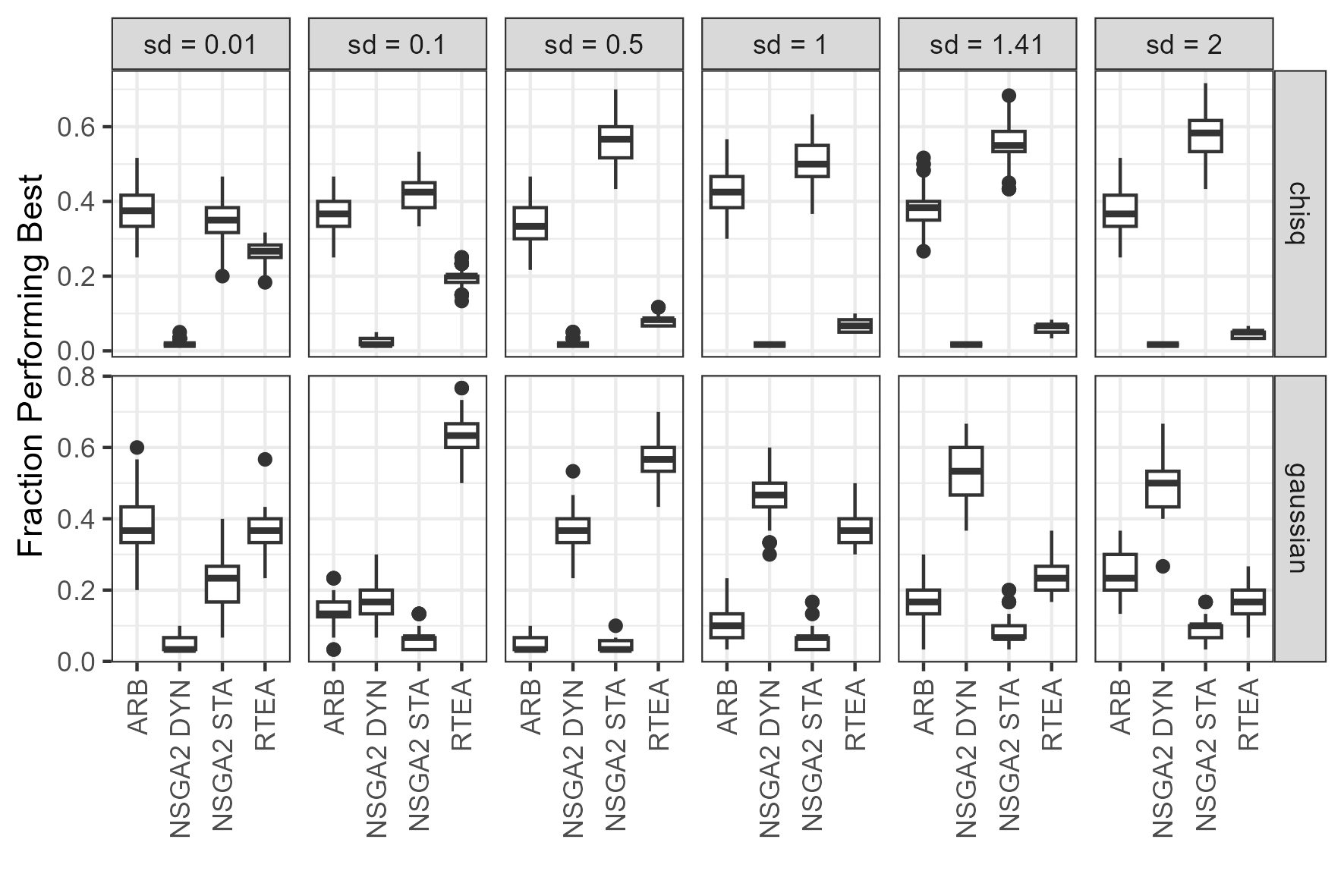}
    \caption{Fraction of performing best of the different clusters of resampling strategies by using just one parameter setting for all problems.}
    \label{fig:Fraction_Best_OverAll}
\end{figure}

\end{document}